\documentclass[a4paper]{article}
\usepackage{comment}
\excludecomment{comment} 
\usepackage[boldmath]{numprint}

\usepackage{times}
\usepackage{latexsym}
\usepackage{graphicx}
\usepackage{todonotes}
\usepackage{amsmath}
\usepackage{color, colortbl}
\definecolor{Gray}{gray}{0.95}
\usepackage[first=0,last=9]{lcg}

\usepackage{amssymb}
\usepackage{stfloats}
\usepackage{float}
\restylefloat{table}
\usepackage{siunitx,booktabs,subcaption,tabularx}
\usepackage{pbox}
\usepackage{hyperref}
\usepackage{comment}
\usepackage{multirow}

\newcommand{\co}{CO$_2$}

\usepackage{INTERSPEECH2022}

\title{Benchmarking Transformers-based models on French Spoken Language Understanding tasks}

\name{Oralie Cattan$^{1,2}$, Sahar Ghannay$^1$, Christophe Servan$^{1,2}$, Sophie Rosset$^1$}
\address{$^1$Universit\'e Paris-Saclay, CNRS, LISN, 91405~Orsay, France \\
$^2$ QWANT, 10 boulevard Haussmann, 75009~Paris, France}
\email{$^1$firstname.lastname@lisn.upsaclay.fr \\ $^2$inital.lastname@qwant.com}
\begin{document}

\maketitle
\begin{abstract}
In the last five years, the rise of the self-attentional Transformer-based architectures led to state-of-the-art performances over many natural language tasks.
Although these approaches are increasingly popular, they require large amounts of data and computational resources. 
There is still a substantial need for benchmarking methodologies ever upwards on under-resourced languages in data-scarce application conditions.
Most pre-trained language models were massively studied using the English language and only a few of them were evaluated on French.
In this paper, we propose a unified benchmark, focused on evaluating models quality and their ecological impact on two well-known French spoken language understanding tasks.
Especially we benchmark thirteen well-established Transformer-based models on the two available spoken language understanding tasks for French: MEDIA and ATIS-FR.
Within this framework, we show that compact models can reach comparable results to bigger ones while their ecological impact is considerably lower. 
However, this assumption is nuanced and depends on the considered compression method.

\end{abstract}

\section{Introduction}
\label{sec:intro}
The development of pre-trained language models based on Transformer architectures \cite{NIPS2017_3f5ee243}, such as BERT \cite{devlin-etal-2019bert} has recently led to significant progress in the field of natural language processing (NLP).
However, the trend of training large pre-trained language models on ever larger corpora, with an ever-increasing amount of parameters has raised questions related to the usability of these approaches \cite{cattan:hal-03336060}.
Because these models require considerable computational resources, major efforts have been made to develop compact models in order to reduce the cost of using them.
Compact models offer alternatives to the energy-intensive models with comparable performances while reducing their computational complexity and size.
These models also allow to solve some industrial problems related to online speech processing.
In particular, some applications (speech recognition, speech to text, etc.) have some known problems associated to network latency, transmission path difficulties, or privacy concerns.

Spoken Language Understanding (SLU) in language dialogue systems refers to the task of producing a semantic analysis and a formalization of the user’s utterance.
SLU traditionally encompasses the processes of determining a broad range of information conveyed in dialogue such as identifying the domain, the intent and the concepts of the conversation.


Benchmarking models have been extensively used in the performance evaluation of NLP-based systems \cite{guo-etal-2020, le-etal-2020-flaubert-unsupervised, abu-farha-magdy-2021-benchmarking}.
With respect to our task of interest, recent research has focused on evaluating word embeddings representations.
Word embeddings have proven to be effective in capturing semantic relationships between words. They are also an essential element of deep learning-based architectures.

In \cite{ghannay-2020-SLUEmbed}, contextual (ELMo \cite{peters-etal-2018deep}) and flat (Word2Vec \cite{DBLP:journals/corr/abs-1301-3781}, GloVe \cite{pennington-etal-2014glove} and Fast-Text \cite{bojanowski-etal-2017enriching}) representations have been evaluated in order to investigate different input representations and their influence on the results obtained in SLU tasks.
They highlighted the competitiveness of Word2Vec and ELMO on the French SLU corpus: MEDIA.
More recently, \cite{ghannay-2020-Coling} investigated the transferability of two French pre-trained BERT models \cite{devlin-etal-2019bert} and their integration in BiLSTM and BiLSTM$+$CNN-based architectures.
They obtained state-of-the-art results on MEDIA using CamemBERT \cite{martin2020camembert} model.
Lately, \cite{multiatis_plusplus} gave a comparison of various cross-lingual transfer approaches including the use of the multilingual BERT encoder, evaluated on MultiATIS++, a multilingual corpus for the task of language understanding.
They show that mBERT brings substantial improvements on multilingual training and cross-lingual transfer tasks, yielding up to 1.4\% of relative improvements on the French subcorpus.

While there are now many English corpora and models for various tasks, there are still very few resources in French. Direct comparisons between SLU models are difficult because of the lack of a unified evaluation framework and size diverse Transformer-based models in French.

\textbf{Contributions}\footnote{This work has been made possible thanks to the Saclay-IA computing platform}: In this study we are investigating the use of transformer-based architectures for French language in solving an SLU problem, a concept detection task~\cite{Maynard06A}.
We also assess the impact related to model compacteness on SLU performance and their ecological impact (e.g.: their \co{} cost).
We focus on the benchmarking of the existing French and multilingual Transformer-based models on two French SLU corpora: MEDIA and ATIS-FR.
Models considered for this benchmark are detailed in section \ref{sec:model_approach}, the experiments, the data, the model optimization and the experimental protocol are described in section \ref{sec:relw}. Section \ref{sec:res} presents the results, and section \ref{sec:analysis} proposes an analysis of some chosen results.

\begin{table*}
\small 
\resizebox{\textwidth}{!}{
\begin{tabular}{lllrlrr}
\hline
\textbf{Model} & \multicolumn{1}{c}{\textbf{Objectives}} & \multicolumn{1}{c}{\textbf{Data}} & \multicolumn{1}{c}{\textbf{Vocabulary size}} & \textbf{Tokenization} & \multicolumn{1}{c}{\textbf{\# parameters}} & \multicolumn{1}{c}{\textbf{Model size}}   \\ \hline \hline
\textbf{FlauBERT}\textsubscript{base}  & MLM & 24 French subcorpora (71 Gb of text)  &  68,729  & BPE       & 138 M                          & 553 Mb                           \\ \hline
\textbf{CamemBERT}\textsubscript{base}  & MLM & French OSCAR (138 Gb of text)      &  32,005 & SentencePiece   & 110 M                          & 445 Mb                           \\
 \textbf{CamemBERT}\textsubscript{base}  & MLM & French CCNet (135 Gb of text)  &  32,005 & SentencePiece   & 110 M                          & 445 Mb                          \\
\textbf{CamemBERT}\textsubscript{base}  & MLM & French OSCAR (4 Gb of text)      &  32,005 & SentencePiece   & 110 M                          & 445 Mb                           \\
\textbf{CamemBERT}\textsubscript{base}  & MLM & French CCNet (4 Gb of text)      &  32,005 & SentencePiece   & 110 M                          & 445 Mb                           \\
\textbf{CamemBERT}\textsubscript{base}  & MLM & French Wikipedia (4 Gb of text) &    32,005 & SentencePiece & 110 M                          & 445 Mb                           \\
 \textbf{CamemBERT}\textsubscript{large}  & MLM & French CCNet (135 Gb of text)  &  32,005    & SentencePiece       & 335 M                          & 1.35 Gb                          \\ \hline
 \textbf{FrALBERT}\textsubscript{base}  & MLM and SOP & French Wikipedia (4 Gb of text) &    32,005  & SentencePiece    & 12 M                           & 50 Mb                            \\ \hline
\textbf{XLM-R}\textsubscript{base}   & MLM & CC-100 (2.5 Tb of text)           &     250,002   & BPE    & 278 M                          & 1.12 Gb                          \\
\textbf{XLM-R}\textsubscript{large}   & MLM & CC-100 (2.5 Tb of text)           &      250,002 & BPE   & 559 M                          & 1.24 Gb                          \\\hline
\textbf{mBERT}\textsubscript{base}   & MLM and NSP & Wiki-100         &     119,547  & WordPiece      & 177 M                          & 714 Mb                        \\
 \textbf{small-mBERT}\textsubscript{base} FR  & MLM and NSP & Wiki-100         &   24,495   & WordPiece      & 104 M                          & 420 Mb                           \\
 \textbf{distil-mBERT}\textsubscript{base}  & MLM and NSP & Wiki-100         &     119,547  & WordPiece      & 134 M                          & 542 Mb                           \\ \hline
\end{tabular}
}
\caption{Model characteristics, in terms of pre-training settings, sources of data and models size.}
\label{table:param}
\end{table*}

\section{Transformer models considered}
\label{sec:model_approach}

In order to compare the models performances on French SLU tasks, we highlight in this section and in Table \ref{table:param} some characteristics of the considered models.


Among the language models that have been proposed in recent years we consider the  most commonly used multilingual benchmarking models that have ``reasonable'' resource consumption: XLM-R \cite{conneau-etal-2020unsupervised} and mBERT \cite{devlin-etal-2019bert}, as well as the French models: CamemBERT \cite{martin2020camembert} and FlauBERT \cite{le-etal-2020-flaubert-unsupervised}~\footnote{We did not include the small CamemBERT models~\cite{micheli-etal-2020importance} and the large version of FlauBERT, the former not being available and the latter not converging in our experiments despite our efforts.} since these are the only two available large models for French at the time of writing.
We also evaluate compact models that are scalable Transformer-based models. We exploit distil-mBERT \cite{Sanh2019DistilBERTAD}, a distilled version of mBERT, small-mBERT \cite{abdaoui-etal-2020load} a mBERT model whose original vocabulary has been reduced to the French language and FrALBERT \cite{cattan:hal-03336060}, a recently released model for French.

As mentioned in \cite{DBLP:journals/corr/abs-2001-08361}, the performance of language models is determined by multiple factors such as the pre-training objective, the layers specification or the dataset size.

Indeed, pre-training objectives vary by model and may depend on the downstream task being solved \cite{10.1162/tacl_a_00300}. 
BERT-like models adopt the masked language modeling (MLM) and next sentence prediction (NSP) objectives.
MLM is a fill-in-the-blank task consisting of predicting tokens of the input sequence that have been masked whereas NSP is a binary classification task to predict whether two segments are adjacent in the original text.
XLM-R differs from mBERT solely in the pre-training procedure, eliminating the NSP task to address its ineffectiveness. 
This is also the case for CamemBERT and FlauBERT models.
FrALBERT pre-training objective consists of MLM and Sentence Order Prediction (SOP) objective.
SOP effectively models inter-sentence coherence by predicting whether a sentence order in a given sentence pair is swapped or not.

For monolingual and multilingual language modeling, the quality and representativeness of the large-scale datasets used are important for efficient modeling and for producing good generalization.
Most models are pre-trained on the content of either Wikipedia, the web pages gathered by Common Crawl (CC) or a combination of diverse corpora.
This represents a few gigabytes to several hundreds or even several terabytes of text.
The other noteworthy difference is the vocabulary size with several tens of thousands of tokens for monolingual models to several hundred for multilingual models.


Models like Text-To-Text Transfer Transformer (T5) \cite{2020t5} or GPT-3 \cite{brown2020language}, have been excluded from our study, since they pose several usability problems (emphasized in the introduction).
Although the GPT-3 and T-5 models are models for English, the data used for their pre-training contains many other languages, however, these languages represent only a small portion of the dataset compared to English (93\% in word count for GPT-3).
Furthermore, in terms of size, GPT-3 and T5 are huge compared to the size of established general NLP models such as BERT, and these earlier models are already quite expensive to run on GPUs.
For example, GPT-3 is trained on billions of parameters, 470 times larger than the BERT model.
We highlight the question of their usability in the introduction section.

In terms of availability finally, BERT is an open-source tool and easily available for users to access and fine-tune according to their needs and solve various downstream tasks. GPT-3 on the other hand is not open-sourced. It has limited access to users.

\section{Experiments}
\label{sec:relw}

Experiments are conducted on two well known French SLU task: MEDIA and ATIS-FR.

\subsection{Datasets}
\label{sec:data}

\textit{The French MEDIA}\footnote{MEDIA is available for academic use:
    \href{https://catalogue.elra.info/en-us/repository/browse/ELRA-S0272/}{\textit{https://catalogue.elra.info/ELRA-S0272}}.} corpus, composed of 1258 transcribed dialogues, which is about hotel reservation and information~\cite{Maynard06A}.
The corpus was manually annotated with semantic concepts characterized by a label and its value.
There are in total $76$ semantic labels.
The corpus is split into three parts: a training corpus composed of 13k sentences, a development corpus composed of 1.3k sentences, and a test corpus composed of 3.5k sentences.
The MEDIA task is also known as one of the most challenging slot-filling tasks, according to \cite{bechet2019benchmarking}.

\textit{The French version of the Air Travel Information System (ATIS)} corpus from the recent MultiATIS++ extension~\cite{multiatis_plusplus}, named ATIS-FR, concerns flight information~\cite{Price:1990:ESL:116580.116612}.
The ATIS-FR corpus corresponds to the manual translation of the original ATIS sentences (in English).
It is composed of $84$ semantic labels, and is split into three parts: a training corpus composed of 4.5k sentences, a development corpus composed of 490 sentences, and a test corpus composed of 893 sentences.

\subsection{Experimental protocol}
\label{sec:data}

Spoken language understanding (SLU) is considered here as a sequence labeling task that assigns a concept label (from a predefined set) to each token of a sentence.
We follow BIO-tagging scheme, where each concept is associated with two labels, B-label (for Beginning) and I-label (for Intermediate). Finally, the O-label (for Other) identifies the non-concept tokens. 
SLU performance is evaluated in terms of F-measure or F1 and Concept Error Rate (CER).
The CER score is the official metric used in the MEDIA campaign \cite{Maynard06A} which is estimated in the same way as the classical word error rate but applied to semantic concepts instead of words (the lower the better). 
The significant results are marked with a star and measured using the 95\% confidence interval.

We rely in this study on the standard approach introduced in \cite{devlin-etal-2019bert}, that corresponds to a model-based transfer learning method, used to facilitate the modeling of the target task with the knowledge learned from the language modeling task.
Specifically, it consists of adding on top of the pre-trained model a token-level classifier with two hidden linear layers (with ReLU activations and dropout) that takes as input the last hidden state of the input sequence and outputs probabilities over concepts.
In our experiments, we use Adam optimizer and conduct an automated hyperparameter optimization on a development dataset.

This optimization is based on the population-based learning algorithm \cite{jaderberg2017population}, with proven efficiency, and in which a population of models and their hyperparameters are jointly optimised. 
Among the hyperparameters considered are the number of training epochs from 5 to 100, the batch size in the interval of 8 and 32 or the learning rate in the range between 1 and 5.

\section{Results}
\label{sec:res}

This section presents the evaluation results of the different Transformer-based pre-trained models described in section~\ref{sec:model_approach} on both French datasets MEDIA and ATIS-FR.

\subsection{MEDIA results}

Performances on the MEDIA test dataset are presented in the two first columns of the Table \ref{table:results}.
Monolingual models scores are very close and vary from 89 to 90 of F1 while CER varies between 7.5 and 8.6.
FlauBERT\textsubscript{base} gets the worst F1 score (89.0) while CamemBERT\textsubscript{base, Wiki 4 Gb}, pre-trained only on 4 Gb of Wikipedia, gets the best F1 score (90.0) with a CER score of 8.4.
Considering the CER, as the official metric of the MEDIA task, the best model is  CamemBERT\textsubscript{base, CCNet 135 Gb}, a base CamemBERT model pre-trained with CCNet on 135 Gb of text. 
It obtains the lowest CER, at 7.5 for an F1 at 89.9.

The multilingual models (mBERT\textsubscript{base}, distill-mBERT\textsubscript{base}, small-mBERT\textsubscript{base-fr}, XLM-R\textsubscript{base} and XLM-R\textsubscript{large}) obtained CER scores ranging from 10.1 to 8.0, for the worst one, distill-mBERT\textsubscript{base}, and the best one XLM-R\textsubscript{large}, respectively.
Regarding the F1 scores, they are comparable to the French monolingual models, for XLM-R models with 89.5 for the \textit{base} one and 89.9 for the \textit{large} one.
On the other side, multilingual BERT and its \textit{distilled} and \textit{small} versions obtained performances comparable to the FlauBERT\textsubscript{base}, with 88.9 F1 at best.

\subsection{ATIS-FR results}

The performance of the monolingual models on the ATIS-FR task in terms of F1 varies between 92.5 and 94.1 and between 3.3 and 5.3 of CER (Table \ref{table:results}). 
FlauBERT gets the worst F1 and CER scores while the large version of the CamemBERT model gets the best F1 and CER scores.
FrALBERT and CamemBERT\textsubscript{base, Wiki 4 Gb} obtain similar performance with 0.3 points of F1 and 0.1 point of CER difference. 
The best model is CamemBERT\textsubscript{large, CCNet 135 Gb}, which obtains the lowest CER, at 3.3 for an F1 at 94.1.
The F1 scores vary from 88.1 (CER at 6.0) for the \textit{distilled} version of mBERT to 93.6 (CER at 5.0) for mBERT\textsubscript{base}. 
The two versions of XLM-R obtained comparable results in terms of CER and F1 scores to CamemBERT\textsubscript{base, Wiki 4 Gb} and FrALBERT\textsubscript{base, Wiki 4 Gb}.

\begin{table}[h!]
\centering
\resizebox{0.48\textwidth}{!}{%
\begin{tabular}{lrrrr}
\hline
& \multicolumn{2}{c}{\textbf{MEDIA}} & \multicolumn{2}{c}{\textbf{ATIS-FR}} \\ 
\textbf{Model}  & \textbf{F1} &\textbf{CER} & \textbf{F1} &\textbf{CER} \\
\hline
\hline
FlauBERT\textsubscript{base} & 89.0 & 8.1 & 92.5 & $^{*}$5.3 \\
\hline
CamemBERT\textsubscript{large, CCNet 135 Gb} & 89.2 & 7.8 & \bf94.1 & \bf$^{*}$3.3 \\
CamemBERT\textsubscript{base, CCNet 135 Gb} &          89.9   &   \bf$^{*}$7.5 & 94.0 & 3.7 \\
CamemBERT\textsubscript{base, OSCAR 138 Gb} &      89.3        &     7.9  & 93.9 & 3.7 \\
CamemBERT\textsubscript{base, OSCAR 4 Gb} & 89.7 & 8.3 & 93.6 &   3.7 \\
CamemBERT\textsubscript{base, CCNet 4 Gb} & 89.7 & 8.3 & 93.8 & 3.8\\
CamemBERT\textsubscript{base, Wiki 4 Gb} & $^{*}$\textbf{90.0} & 8.4 & 92.5 & 4.2 \\
\hline
FrALBERT\textsubscript{base, Wiki 4 Gb} & 89.8 & 8.6 & 92.8 & 4.3 \\
\hline
XLM-R\textsubscript{base} & 89.5 & 8.5 & 92.5 & 4.3 \\
XLM-R\textsubscript{large} & 89.9 & 8.0 & 92.7 & 4.4 \\
\hline
mBERT\textsubscript{base} & 88.9 & 8.7 &  93.6 &  5.0 \\
distill-mBERT\textsubscript{base} & $^{*}$87.5 & $^{*}$10.1 & $^{*}$88.1 & $^{*}$6.0 \\
small-mBERT\textsubscript{base-fr}  & $^{*}$88.8 & $^{*}$8.1 &  93.3 & $^{*}$5.3 \\
\hline
\end{tabular}
}
\caption{SLU performances on the MEDIA and ATIS-FR test dataset. Results are given in terms of F-measure (F1) and Concept Error Rate (CER). Significant results are marked with a star.}
\label{table:results}
\end{table}

\vspace{-0.7cm}
\section{Analysis}
\label{sec:analysis}

\subsection{In deph analysis}




Benchmarking Transformer-based models on the two French SLU tasks (MEDIA \& ATIS-FR) allows to observe some trends.
In both tasks, the CamemBERT models perform the best, in terms of Concept Error Rate (CER).
FrALBERT obtains comparable results to CamemBERT\textsubscript{base, Wiki 4 Gb}, this may come from that the two models are trained on the same kind and amount of data (Wikipedia 4G).
We also observe that FrALBERT has comparable performances to XLM-R\textsubscript{base} in both tasks even if the training data and structure are both different.
In addition, we notice the underachievement of distill-mBERT\textsubscript{base} compared to other BERT models and especially to FrALBERT.
Finally, FlauBERT\textsubscript{base} has comparable results to CamemBERT models in the MEDIA task, but the CER score of FlauBERT in the ATIS-FR task, is significantly worse than the CamemBERT models,  which led us to go deeper in the analysis.


We propose to focus the analysis on the most representative labels of each task. 
For MEDIA we focus on \textsc{command-tache}, \textsc{temps-date}, \textsc{nombre-chambre}, \textsc{localisation-ville}, \textsc{nom-hotel}, \textsc{objet}, labels. 
Their F1 varies between 70.15 and 95.76.
In the ATIS task, we focus on \textsc{city\_name}, \textsc{airline\_name}, \textsc{depart\_time.period\_of\_day},\textsc{depart\_date.day\_name}, \textsc{toloc.city\_name}, \textsc{fromloc.city\_name} labels and their F1 varies between 59.41 and 100.


The first trend we observed is an underperformance of the distill-mBERT\textsubscript{base} on named entity tags on both tasks.
For instance, \textsc{state name}, \textsc{city\_name}  in ATIS or \textsc{nom-hotel}, \textsc{localisation-ville} for MEDIA have up to 2 F1 points less than the other models. 
Note that these tags are ones of the most frequent in both tasks. 
On the other side, we could not observe such a big trend by comparing larger models and compact models, or multilingual models versus French monolingual models.
Differences between models are very small, for instance the CamemBERT\textsubscript{base, Wiki 4 Gb} model will be better than the FrALBERT\textsubscript{base, Wiki 4G} on the MEDIA labels  
\textsc{nombre-chambre} (93,14 versus  91,85 of F1 respectively), \textsc{nom-hotel} (82,76 versus  79,60 of F1 respectively).


When diving deeper in the MEDIA task, we can detect some small trends when we look at the whole results.
For instance, even if the \textsc{objet} label is one of the most frequent, it seems that all models have difficulties to correctly detect it.
In the same way, the label \textsc{nom} is highly difficult, even for one of the best model (CamemBERT\textsubscript{base, CCNet 135 Gb}). 
It seems that the biggest difference between models occurs in their ability to transfer their knowledge according to the amount of data used for pretraining.
In this way, the models trained with the biggest corpora are enable to handle the best, the most infrequent labels.

What we observe in the MEDIA task can also be observed in the ATIS-FR task, even if the overall performance of the models is higher. 
This leads us to suggest that a better sample of the training data could bring interesting results.
Moreover, recent works in few-shot learning for slot-filling approaches \cite{cattan2021metanlp} enable models to perform at high levels with a strong sub-sampling of examples, which in this case could be a plus.







\npdecimalsign{.}
\nprounddigits{3}
\begin{table*}[ht!]
\centering
\resizebox{\textwidth}{!}{%
\begin{tabular}{l|rrrrrr|rrrrrr}
\hline
\textbf{Steps}& \multicolumn{6}{c}{\textbf{Fine-tuning \textit{(1 epoch)}}} & \multicolumn{6}{|c}{\textbf{Inference}} \\ 
\textbf{Tasks}&  \multicolumn{3}{c}{\textbf{MEDIA}} & \multicolumn{3}{c}{\textbf{ATIS-FR}} & \multicolumn{3}{|c}{\textbf{MEDIA}} & \multicolumn{3}{c}{\textbf{ATIS-FR}} \\ 
\hline
\multirow{2}{*}{\textbf{Models}}  & \textbf{Time} &\textbf{Energy} &\textbf{\co{}} &\textbf{Time} &\textbf{Energy} &\textbf{\co{}} & \textbf{Time} &\textbf{Energy} &\textbf{\co{}} &\textbf{Time} &\textbf{Energy} &\textbf{\co{}} \\
  & \textbf{(s)} &\textbf{(kWh)} &\textbf{(g)} &\textbf{(s)} &\textbf{(kWh)} &\textbf{(g)} & \textbf{(s)} &\textbf{(kWh)} &\textbf{(g)} &\textbf{(s)} &\textbf{(kWh)} &\textbf{(g)} \\
\hline
\hline
FlauBERT\textsubscript{base} & \nprounddigits{2}$\numprint{121.89086890220642}$  & \nprounddigits{2}$\numprint{765.2435453088966}$  & \nprounddigits{2}$\numprint{554.0363268036411}$  & \nprounddigits{2}$\numprint{52.07612047195435}$ & \nprounddigits{2}$\numprint{231.13451947199545}$  & \nprounddigits{2}$\numprint{167.34139209772474}$ & \nprounddigits{2}$\numprint{9.438830852508543}$   & \nprounddigits{2}$\numprint{26.52477602853584}$ & \nprounddigits{2}$\numprint{19.20393784465995}$ & \nprounddigits{2}$\numprint{1.4418760299682618}$ & \nprounddigits{2}$\numprint{4.007720076605968}$ & \nprounddigits{2}$\numprint{2.90158933546272}$ \\
\hline
CamemBERT\textsubscript{large, CCNet 135 Gb} & \nprounddigits{2}$\numprint{144.3085873126984}$  & \nprounddigits{2}$\numprint{659.335882559118}$  & \nprounddigits{2}$\numprint{477.3591789728014}$  & \nprounddigits{2}$\numprint{56.64867458343506}$ & \nprounddigits{2}$\numprint{345.3262290298859}$  & \nprounddigits{2}$\numprint{250.01618981763747}$ & \nprounddigits{2}$\numprint{23.665659725666046}$   & \nprounddigits{2}$\numprint{66.57963756906129}$ & \nprounddigits{2}$\numprint{48.203657600000376}$ & \nprounddigits{2}$\numprint{3.442496347427368}$  & \nprounddigits{2}$\numprint{9.636250469627835}$ & \nprounddigits{2}$\numprint{6.976645340010551}$  \\
CamemBERT\textsubscript{base, OSCAR 138 Gb} & \nprounddigits{2}$\numprint{130.055819272995}$  & \nprounddigits{2}$\numprint{789.6946701255281}$  & \nprounddigits{2}$\numprint{571.7389411708824}$  & \nprounddigits{2}$\numprint{53.39182577133179}$  & \nprounddigits{2}$\numprint{234.83683421287225}$  & \nprounddigits{2}$\numprint{170.0218679701195}$ & \nprounddigits{2}$\numprint{9.459187388420105}$   & \nprounddigits{2}$\numprint{26.58016572490125}$ & \nprounddigits{2}$\numprint{19.244039984828508}$ & \nprounddigits{2}$\numprint{1.5256008148193358}$ & \nprounddigits{2}$\numprint{4.23319313437604}$ & \nprounddigits{2}$\numprint{3.064831829288253}$ \\
CamemBERT\textsubscript{base, CCNet 135 Gb} & \nprounddigits{2}$\numprint{118.57637429237366}$  & \nprounddigits{2}$\numprint{671.4163189787637}$  & \nprounddigits{2}$\numprint{486.10541494062494}$  & \nprounddigits{2}$\numprint{51.6749819755554}$ & \nprounddigits{2}$\numprint{230.00736052588633}$  & \nprounddigits{2}$\numprint{166.52532902074168}$ & \nprounddigits{2}$\numprint{7.279335403442383}$  & \nprounddigits{2}$\numprint{20.443919048920133}$ & \nprounddigits{2}$\numprint{14.801397391418176}$ & \nprounddigits{2}$\numprint{1.18954439163208}$  & \nprounddigits{2}$\numprint{3.223501276225785}$ & \nprounddigits{2}$\numprint{2.333814923987469}$ \\
CamemBERT\textsubscript{base, OSCAR 4 Gb} & \nprounddigits{2}$\numprint{116.59348160028458}$   & \nprounddigits{2}$\numprint{623.4403681269781}$  & \nprounddigits{2}$\numprint{451.37082652393207}$  & \nprounddigits{2}$\numprint{51.66041407585144}$  & \nprounddigits{2}$\numprint{229.96171820790943}$  & \nprounddigits{2}$\numprint{166.49228398252637}$ & \nprounddigits{2}$\numprint{7.4292685985565186}$  & \nprounddigits{2}$\numprint{20.865448213804687}$ & \nprounddigits{2}$\numprint{15.106584506794594}$ & \nprounddigits{2}$\numprint{1.1835417747497559}$ & \nprounddigits{2}$\numprint{3.272318959599463}$ & \nprounddigits{2}$\numprint{2.369158926750011}$ \\
CamemBERT\textsubscript{base, CCNet 4 Gb} & \nprounddigits{2}$\numprint{115.54236489534378}$  & \nprounddigits{2}$\numprint{662.7838950202784}$  & \nprounddigits{2}$\numprint{479.85553999468146}$  & \nprounddigits{2}$\numprint{50.735571670532224}$  & \nprounddigits{2}$\numprint{227.3475416878081}$  & \nprounddigits{2}$\numprint{164.59962018197308}$ & \nprounddigits{2}$\numprint{7.308168935775757}$  & \nprounddigits{2}$\numprint{20.52596692168711}$ & \nprounddigits{2}$\numprint{14.860800051301467}$ & \nprounddigits{2}$\numprint{1.2076034545898438}$ & \nprounddigits{2}$\numprint{3.309590093361817}$ & 	\nprounddigits{2}$\numprint{2.3961432275939554}$ \\
CamemBERT\textsubscript{base, Wiki 4 Gb}  & \nprounddigits{2}$\numprint{109.56707954406738}$  & \nprounddigits{2}$\numprint{645.9618365370154}$  & \nprounddigits{2}$\numprint{467.6763696527991}$  & \nprounddigits{2}$\numprint{50.395973443984985}$  & \nprounddigits{2}$\numprint{226.39482086850103}$  & \nprounddigits{2}$\numprint{163.90985030879477}$ & \nprounddigits{2}$\numprint{7.190913486480713}$   & \nprounddigits{2}$\numprint{20.19315101332504}$ & \nprounddigits{2}$\numprint{14.619841333647328}$ & \nprounddigits{2}$\numprint{1.1721922874450683}$ & \nprounddigits{2}$\numprint{3.2395157164059535}$ & \nprounddigits{2}$\numprint{2.34540937867791}$ \\
\hline
FrALBERT\textsubscript{base, Wiki 4 Gb} & \nprounddigits{2}$\numprint{72.65081276893616}$  & \nprounddigits{2}$\numprint{474.6854391208822}$  & \nprounddigits{2}$\numprint{343.6722579235187}$  & \nprounddigits{2}$\numprint{28.551519012451173}$  & \nprounddigits{2}$\numprint{97.2550686584902}$  & \nprounddigits{2}$\numprint{70.4126697087469}$ & \nprounddigits{2}$\numprint{4.262052893638611}$  & \nprounddigits{2}$\numprint{11.949288561378564}$ & \nprounddigits{2}$\numprint{8.65128491843808}$ & 	\nprounddigits{2}$\numprint{0.6500789642333984}$  & 
\nprounddigits{2}$\numprint{1.7785563899893126}$ & \nprounddigits{2}$\numprint{1.2876748263522625}$ \\
\hline
XLM-R\textsubscript{base} & \nprounddigits{2}$\numprint{125.25779747962953}$   & \nprounddigits{2}$\numprint{549.475609894469}$ & \nprounddigits{2}$\numprint{397.8203415635955}$ & \nprounddigits{2}$\numprint{56.29638571739197}$  & \nprounddigits{2}$\numprint{243.01747202764844}$  & \nprounddigits{2}$\numprint{175.9446497480174}$ & \nprounddigits{2}$\numprint{8.04273372888565}$ & \nprounddigits{2}$\numprint{22.58923831716039}$ & \nprounddigits{2}$\numprint{16.35460854162412}$ & 	\nprounddigits{2}$\numprint{1.207436990737915}$  & 
\nprounddigits{2}$\numprint{3.339681600784099}$ & \nprounddigits{2}$\numprint{2.4179294789676875}$ \\
XLM-R\textsubscript{large} & \nprounddigits{2}$\numprint{196.73727932572365}$ & \nprounddigits{2}$\numprint{1155.149284704345}$ & \nprounddigits{2}$\numprint{836.3280821259457}$ & \nprounddigits{2}$\numprint{64.08204770088196}$ & \nprounddigits{2}$\numprint{433.70149539807295}$ & \nprounddigits{2}$\numprint{313.9998826682048}$ & \nprounddigits{2}$\numprint{26.027471721172333}$  & \nprounddigits{2}$\numprint{73.23644475143703}$ & \nprounddigits{2}$\numprint{53.0231860000404}$ & \nprounddigits{2}$\numprint{3.7505123019218445}$  & \nprounddigits{2}$\numprint{9.881490306456148}$ & \nprounddigits{2}$\numprint{7.60530165414247}$ \\
\hline
mBERT\textsubscript{base} & \nprounddigits{2}$\numprint{119.36135707582746}$ & \nprounddigits{2}$\numprint{673.560808293512}$  & \nprounddigits{2}$\numprint{487.65802520450274}$  & \nprounddigits{2}$\numprint{52.60944724082947}$ & \nprounddigits{2}$\numprint{232.64777118717552}$  & \nprounddigits{2}$\numprint{168.436986339515}$ & \nprounddigits{2}$\numprint{8.387088418006897}$   & \nprounddigits{2}$\numprint{23.559137365578838}$ & \nprounddigits{2}$\numprint{17.056815452679082}$ & 	\nprounddigits{2}$\numprint{1.1590777397155763}$ &   \nprounddigits{2}$\numprint{3.213201327332255}$ & \nprounddigits{2}$\numprint{2.326357760988553}$ \\
distill-mBERT\textsubscript{base} & \nprounddigits{2}$\numprint{80.10337328910829}$  & \nprounddigits{2}$\numprint{545.4623003043376}$  & \nprounddigits{2}$\numprint{394.9147054203404}$  & \nprounddigits{2}$\numprint{49.48211336135864}$  & \nprounddigits{2}$\numprint{240.9624332941549}$  & \nprounddigits{2}$\numprint{166.52532902074168}$ & \nprounddigits{2}$\numprint{7.03565239906311}$   & \nprounddigits{2}$\numprint{19.752404897113156}$ & \nprounddigits{2}$\numprint{14.300741145509923}$ & \nprounddigits{2}$\numprint{1.0990451335906983}$  & \nprounddigits{2}$\numprint{3.0166680247884994}$ & \nprounddigits{2}$\numprint{2.1840676499468732}$ \\
small-mBERT\textsubscript{base-fr}  & \nprounddigits{2}$\numprint{112.08162021636963}$ & \nprounddigits{2}$\numprint{589.8396314377551}$  & \nprounddigits{2}$\numprint{427.04389316093466}$  & \nprounddigits{2}$\numprint{50.671253859996796}$ & \nprounddigits{2}$\numprint{227.16688289555267}$  & \nprounddigits{2}$\numprint{164.46882321638017}$ & \nprounddigits{2}$\numprint{7.692869472503662}$    & \nprounddigits{2}$\numprint{21.593639575437955}$  & \nprounddigits{2}$\numprint{15.633795052617078}$  & 	\nprounddigits{2}$\numprint{1.152106475830078}$  & \nprounddigits{2}$\numprint{3.190482275707969}$ & \nprounddigits{2}$\numprint{2.30990916761257}$   \\
\hline
\end{tabular}
}
\caption{Estimation of ﬁne-tuning and inference costs on MEDIA and ATIS-FR corpora.}
\label{table:ecoresults}
\end{table*}
\npnoround

\subsection{Ecological and computational costs}


We conducted an impact study of Transformer models we considered by measuring the ecological impact of these models \cite{henderson2020towards}. 
The table \ref{table:ecoresults} presents the time to process one fine-tuning epoch, and one inference step, associated to the energy consumed and the \co{} produced, for each task. 
Each score is the mean of five runs for each step.


CamemBERT\textsubscript{large, CCNet 135 Gb} and XLM-R\textsubscript{large} produce the most over all tasks and steps. 
FlauBERT\textsubscript{base}, CamemBERT\textsubscript{base, CCNet 135 Gb} and mBERT\textsubscript{base} produce nearly the same amount of \co{} during fine-tuning and inference steps and for the both tasks (MEDIA and ATIS-FR), which lead us to correlate the amount of parameters and their impact in terms of  energy consumed and \co{} produced.

FrALBERT\textsubscript{base, Wiki 4 Gb} is the model which uses the less energy (in both fine-tuning and inference steps), and produces the less \co{} of all models. 
Especially, in the inference step, the FrALBERT model consumes nearly the half of other compact models: distill-mBERT\textsubscript{base} and small-mBERT\textsubscript{base-fr}. 

This study shows compact models like FrALBERT, which obtained comparable results to bigger models (especially with large BERT models in Table \ref{table:results}), have also an important lower ecological impact than big Transformer models. 
For instance, the impact of FrALBERT is more than 5 time lower than the impact of large models such as XML-R\textsubscript{large} and CamemBERT\textsubscript{large} in the inference step for both tasks and in the fine-tuning step, the FrALBERT model is more than 3 times lower than the XML-R\textsubscript{large} model.

Surprisingly, the compression approaches of compact models may not have the same impact. 
We can observe that distill-mBERT\textsubscript{base} is a little bit different from small-mBERT\textsubscript{base} with a resource requirement and thus an impact a little bit lower but still quite comparable to the mBERT\textsubscript{base} model, but with significantly lower performances, compared to mBERT.


\section{Conclusion}

Transformer-based architectures are currently the state-of-the-art model for many NLP tasks.
In this study we have proposed to benchmark the existing French and multilingual Transformer-based pre-trained models, for the purpose of comparing their performance on two French SLU corpora: MEDIA and ATIS-FR.
We have also assessed the ecological impact related to model compacteness on SLU performances and conducted an extensive side-by-side comparison of thirteen recently proposed efficient Transformer models.

The experimental results show that these tasks are very challenging even for large models.
Experimental results show that for both tasks, the CamemBERT models perform the best, in term of Concept Error Rate (CER).
The best CER results on MEDIA and ATIS-FR are respectively achieved by CamemBERT\textsubscript{base, CCNet 135 Gb} and CamemBERT\textsubscript{large, CCNet 135 Gb}.
In both tasks, the French compact model FrALBERT obtains comparable results to the large model CamemBERT (\textit{base} and \textit{Wiki 4 Gb} configuration).
Moreover, this model achieves comparable performances to multilingual models in both tasks and outperforms the distilled version.

Then, our detailed analysis of F1 scores provides interesting model insights in each task.
From those analyses we observe that the biggest difference between models occurs in their ability to transfer their knowledge according to the amount of pre-trained data. 

Finally, the ecological study conducted on these models shows that compact models have  significantly less ecological impact compared to big Transformer models in both fine-tuning and inference steps.
This less \co{} is a plus in a context of reducing our ecological impact but it also means less energy consumed, which can reach more than 5 time less between the FrALBERT model and large BERT models. 


We plan to open source our code and benchmarks to facilitate future benchmarking, research and model development.


\bibliographystyle{IEEEtran}

\bibliography{strings,anthology}

\end{document}